\documentclass{article}
\usepackage{spconf,amsmath,epsfig}
\graphicspath{{Figures/}}
\usepackage{graphicx}
\usepackage{lipsum}
\usepackage{url}
\usepackage{color}
\usepackage{bm}
\usepackage{balance}
\usepackage{tabto}

\usepackage{array}
\newcommand{\PreserveBackslash}[1]{\let\temp=\\#1\let\\=\temp}
\newcolumntype{C}[1]{>{\PreserveBackslash\centering}p{#1}}
\newcolumntype{R}[1]{>{\PreserveBackslash\raggedleft}p{#1}}
\newcolumntype{L}[1]{>{\PreserveBackslash\raggedright}p{#1}}

\title{Using GAN\lowercase{s} to Augment Data for Cloud Image Segmentation Task}
\name{Mayank Jain$^{1,2}$, Conor Meegan$^{2}$, and Soumyabrata Dev$^{1,2}$
\thanks{The ADAPT Centre for Digital Content Technology is funded under the SFI Research Centres Programme (Grant 13/RC/2106) and is co-funded under the European Regional Development Fund.}
\thanks{Send correspondence to S.\ Dev: \protect\url{soumyabrata.dev@ucd.ie}}
}
\address{$^{1}$The ADAPT SFI Research Centre, Dublin, Ireland \\
$^{2}$School of Computer Science, University College Dublin, Ireland}

\begin{document}

\maketitle
\begin{abstract}
While cloud/sky image segmentation has extensive real-world applications, a large amount of labelled data is needed to train a highly accurate models to perform the task. Scarcity of such volumes of cloud/sky images with corresponding ground-truth binary maps makes it highly difficult to train such complex image segmentation models. In this paper, we demonstrate the effectiveness of using Generative Adversarial Networks (GANs) to generate data to augment the training set in order to increase the prediction accuracy of image segmentation model. We further present a way to estimate ground-truth binary maps for the GAN-generated images to facilitate their effective use as augmented images. Finally, we validate our work with different statistical techniques.
\end{abstract}
\section{Introduction}
\label{sec:intro}

Determining cloud coverage over a specific location at a given time plays an important role in forecasting key weather related parameters like rainfall, humidity, and solar irradiance~\cite{ma2018application}. Additionally, exact spread of the clouds has been proven to affect the power generated from the photovoltaic systems~\cite{xiang2017very}. While many a studies for cloud analysis have been done using satellite images, they generally suffer from low temporal and/or spatial resolution limiting their utility. This has led to a growing popularity of ground-based sky cameras, also known as Whole-Sky-Imagers~\cite{long2006retrieving,dev2017color}. Although, images captured by these cameras have good temporal resolution with localized focus, they are too noisy and have limited information making them difficult to segment.

Traditional image processing techniques for image segmentation tasks have generally been outperformed by more recent advent of deep learning based methods~\cite{sultana2020evolution}. It is also applicable for the task of sky/cloud image segmentation. Using a total of $1128$ annotated images, Dev~et~al.~\cite{dev2019cloudsegnet} trained a deep convolutional neural network (\textit{CloudSegNet}) and reported a maximum F-score of nearly $0.90$ as compared to less than $0.80$ for previous efforts without deep networks~\cite{long2006retrieving,dev2017nighttime}. Similarly, Xie~et~al.~\cite{xie2020segcloud} reported a pixel-wise classification accuracy of more than $95\%$ after training their proposed architecture called \textit{SegCloud}.

Although deep learning architectures have shown great promise for the task, they need significantly high volumes of labelled data to effectively optimize the large number of parameters and hyperparameters. Scarcity of such data in the case of cloud/sky image segmentation makes the it highly difficult to achieve higher accuracy and achieve robustness. Hence, determining a method to generate such data in an automated way will be a huge help to generate highly accurate cloud/sky image segmentation models.

In recent years, Generative Adversarial Networks (GANs) and its variants have been successfully used to generate synthetic images which look very similar to the real ones~\cite{goodfellow2014generative}. Modified GAN architectures are also suggested in the literature to append class labels as conditions for the GANs to generate images~\cite{antoniou2017data}. While such variants help to generate class labels corresponding to the auto-generated images for the image classification tasks, it is still very difficult to generate ground-truth segmentation maps for the image segmentation tasks.

Using a publicly available dataset of sky/cloud images with corresponding segmentation labels (Section~\ref{sec:dataset}), we train a GAN to automatically generate new sky/cloud images (Section~\ref{sec:GANimageGen}). Ground truth segmentation labels are then estimated by an unsupervised clustering algorithm (Section~\ref{sec:gt_estimation}). A simple regression model is then trained on the both non-augmented set (Section~\ref{sec:PLS}) and the augmented set (Section~\ref{sec:gan_augmentation}) to draw a meaningful comparison. The obtained results are then discussed in Section~\ref{sec:results} and the paper is finally concluded in Section~\ref{sec:conclusion}. 

\section{Dataset}\label{sec:dataset}
In this study, we used a relatively small dataset of night-time images called SWINSEG dataset~\cite{dev2017nighttime}. The dataset contains a total of $115$ images (with $500\times500$ pixel resolution) depicting night-time cloud/sky patterns. Furthermore, the ground-truth binary segmentation maps are also provided in the dataset. 

\begin{figure}[!ht]
    \centering
    \includegraphics[trim={0 302 0 0},clip,width=0.75\columnwidth]{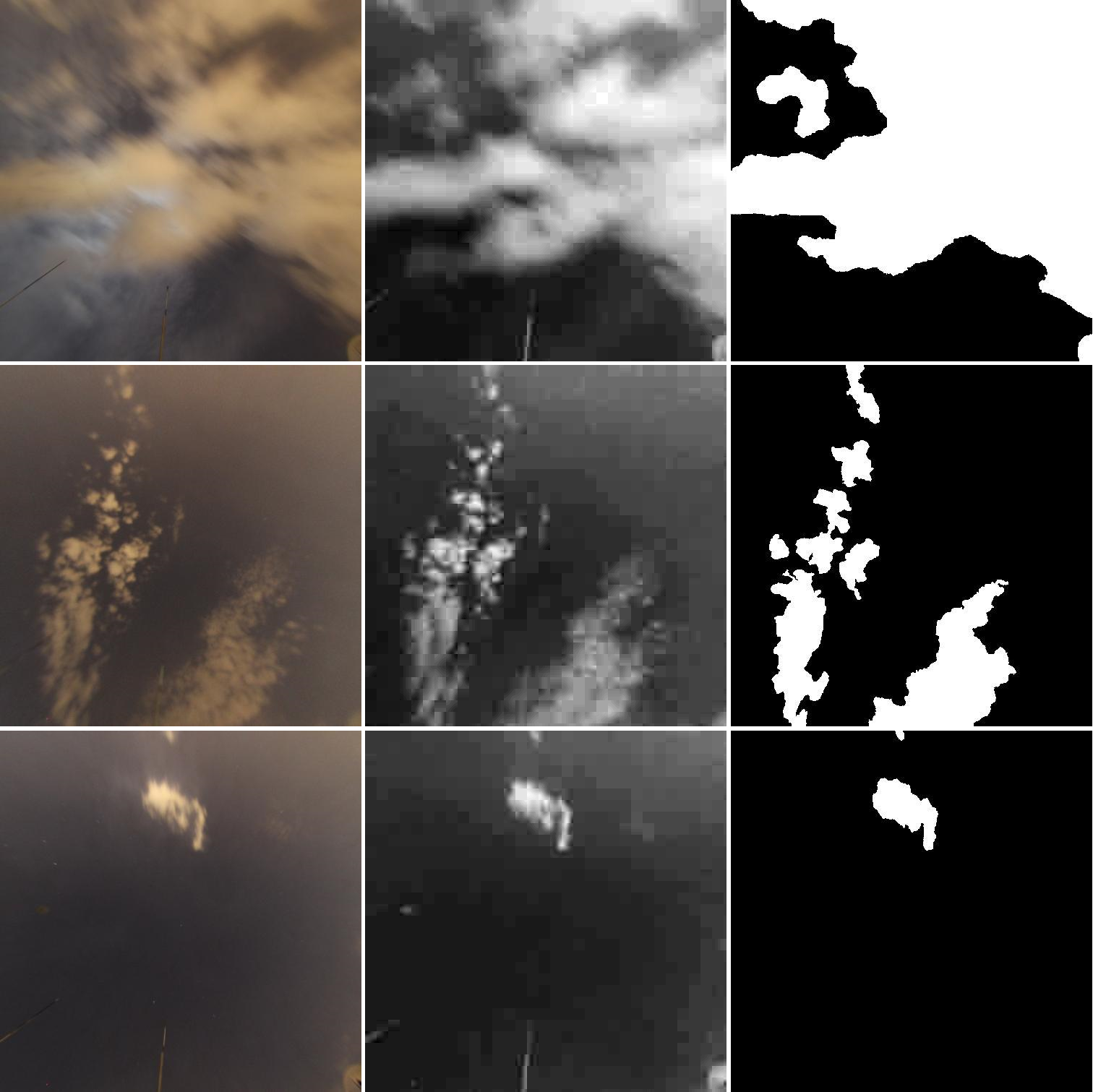}
    \caption{Sample images from the used SWINSEG dataset. \textit{First column:} original $RGB$ images. \textit{Second column:} extracted $R-B$ channel. \textit{Last column:} corresponding ground-truth binary segmentation maps.}
    \label{fig:datasetwithRBchannel}
\end{figure}

One of the major difficulties while dealing with the night-time cloud/sky images is the blur which occurs between the edges of sky and clouds. To resolve this issue, we use $R-B$ channel only for our analysis throughout this study~\cite{dev2017nighttime}. Fig.~\ref{fig:datasetwithRBchannel} shows a few images that are available in the dataset along with the extracted $R-B$ channel and the provided ground-truth binary maps.

For the purpose of training the image segmentation model, we have split the given dataset into the training set, the validation set, and the test set. The splitting is as follows:
\begin{itemize}
    \setlength\itemsep{0em}
    \item Training Set:\tabto{3cm} $69$ Images ($60\%$)
    \item Validation Set:\tabto{3cm} $18$ Images ($15.65\%$)
    \item Test Set:\tabto{3cm} $28$ Images ($24.35\%$)
\end{itemize}

\section{Methodology}\label{sec:methodology}
The task is divided into two stages. In the first stage, we train a GAN architecture to generate cloud/sky images, followed by the estimation of the corresponding ground-truth binary maps. The second stage deals with the training of a prediction model to perform the cloud image segmentation. This is done using the Partial Least Squares (PLS) regression model~\cite{wegelin2000asurvey}. 

\subsection{GAN for cloud/sky image generation}\label{sec:GANimageGen}
\begin{figure}[!ht]
    \centering
    \includegraphics[width=0.98\columnwidth]{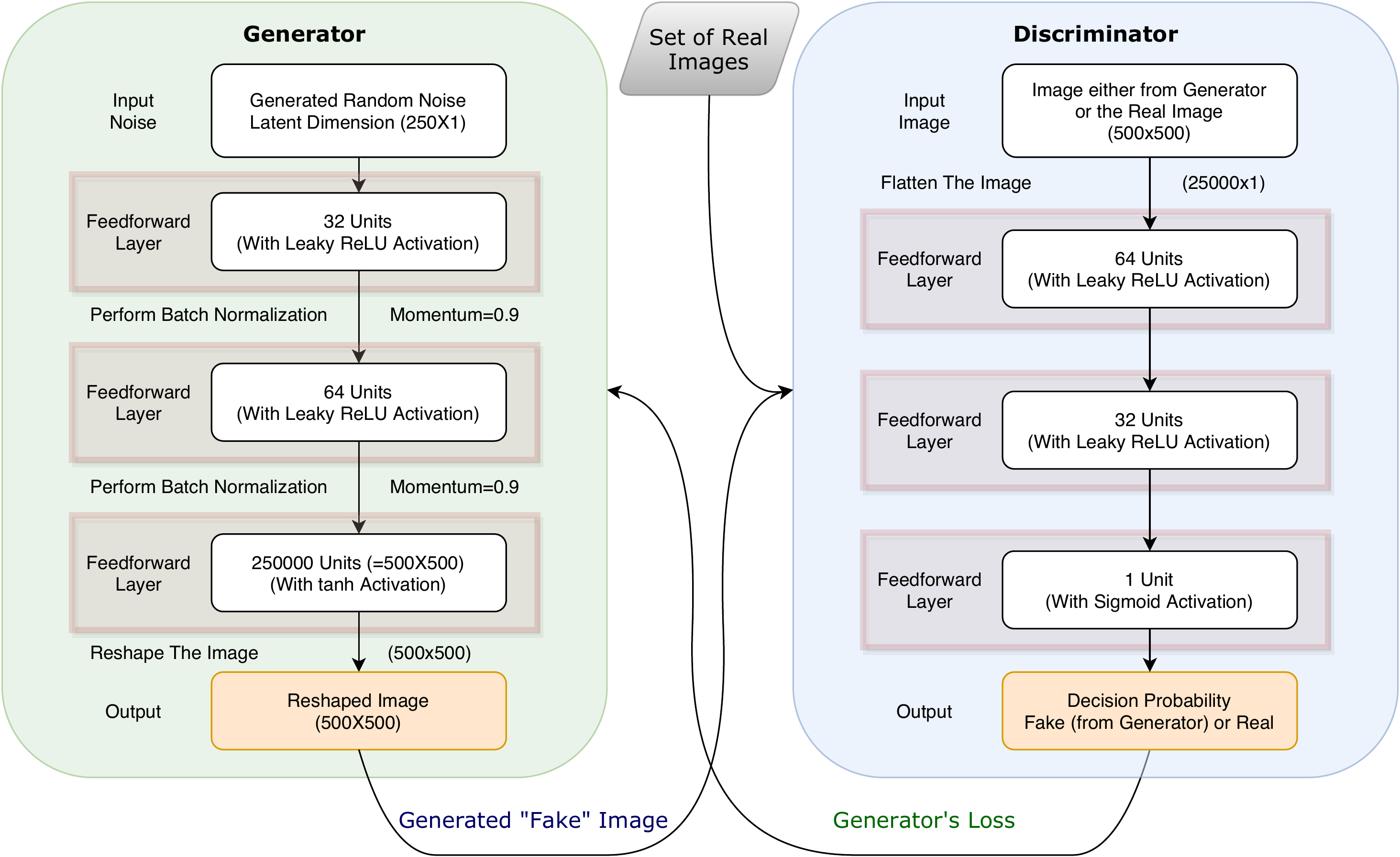}
    \caption{GAN architecture used to generate synthetic cloud/sky images which look similar to the original image.}
    \label{fig:GANarchitecture}
    \vspace{-0.3cm}
\end{figure}

Generative Adversarial Networks (GANs) are highly valued for their ability to generate synthetic images which look similar to the real ones~\cite{goodfellow2014generative}. A GAN is composed of two parts, namely, a generator and a discriminator. While the generator's task is to generate images from latent noise such that the discriminator fails to classify it as `fake', the discriminator's task is to successfully segregate fake images that are generated by the generator from the real images. Fig.~\ref{fig:GANarchitecture} shows the GAN architecture which is used in this study.

Before proceeding with the training of the defined GAN, training data is augmented with basic image transformation procedures of rotation and reflection. The images are thus 16-folded by rotating the original images (by $90^{\circ}$, $180^{\circ}$, and $270^{\circ}$) and then reflecting the resulting images both vertically and horizontally. Post augmentation, the images are normalized in the range $[-1, 1]$ for the training process (see equation~\ref{eq:GANpre-normalization}). This means that the generator also learns to construct images in the same range. Hence, before finally using the generator to generate images for the second stage of this study, we again convert them back to the actual range of $[0, 255]$ (see equation~\ref{eq:GANpost-denormalization}).
\vspace{-0.2cm}
\begin{equation}
    pixel_{[-1, 1]} = \frac{pixel_{[0, 255]}}{127.5} - 1
    \label{eq:GANpre-normalization}
\end{equation}
\vspace{-0.4cm}
\begin{equation}
    pixel_{[0, 255]} = \lfloor (pixel_{[-1, 1]} \times 127.5) + 127.5 \rceil
    \label{eq:GANpost-denormalization}
\end{equation}\vspace{-0.1cm}

\begin{figure}[!ht]
    \centering
    \includegraphics[trim={0 0 0 302},clip,width=0.75\columnwidth]{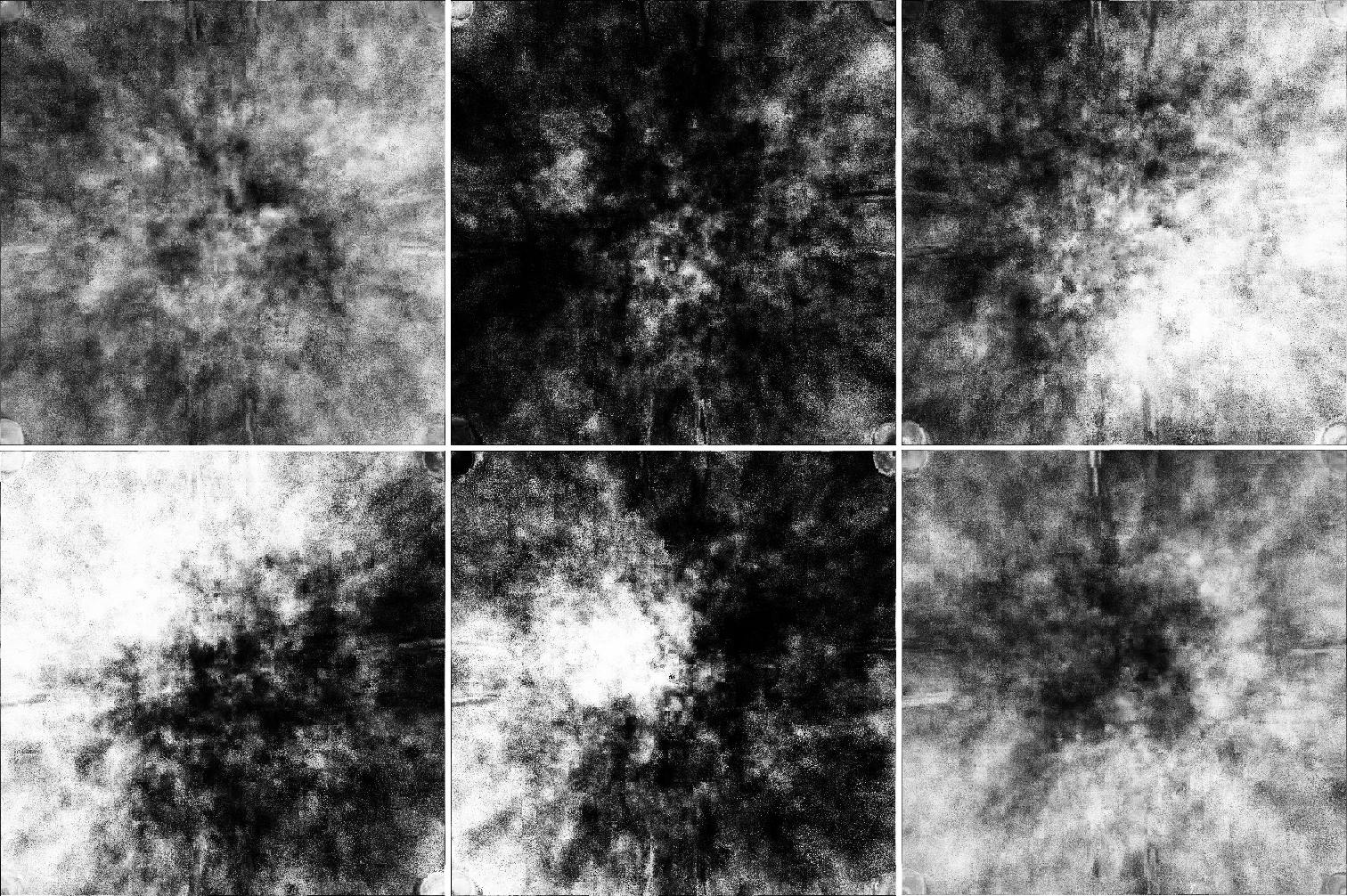}
    \caption{Sample cloud/sky images (in $R-B$ channel) that were generated by the trained GAN.}
    \label{fig:GANgenImages}
    \vspace{-0.25cm}
\end{figure}

The GAN is finally trained using the Adam optimizer~\cite{kingma2014adam} (with a learning rate of $0.00025$) and the cross-entropy loss function with a batch size of $32$ and for $1000$ epochs using the TensorFlow $2.1.0$ library over MX150 GPU (with $2$GB memory) running CUDA $10.1$ and cuDNN $7.6$ version. Fig.~\ref{fig:GANgenImages} shows some of the cloud/sky images that were generated by the trained generator. Since all the original images were converted to $R-B$ channel (as described in Section~\ref{sec:dataset}), the generated images correspond to the processed $R-B$ channel.

\subsubsection{Ground truth estimation}
\label{sec:gt_estimation}
\begin{figure}[!ht]
    \centering
    \includegraphics[width=0.75\columnwidth]{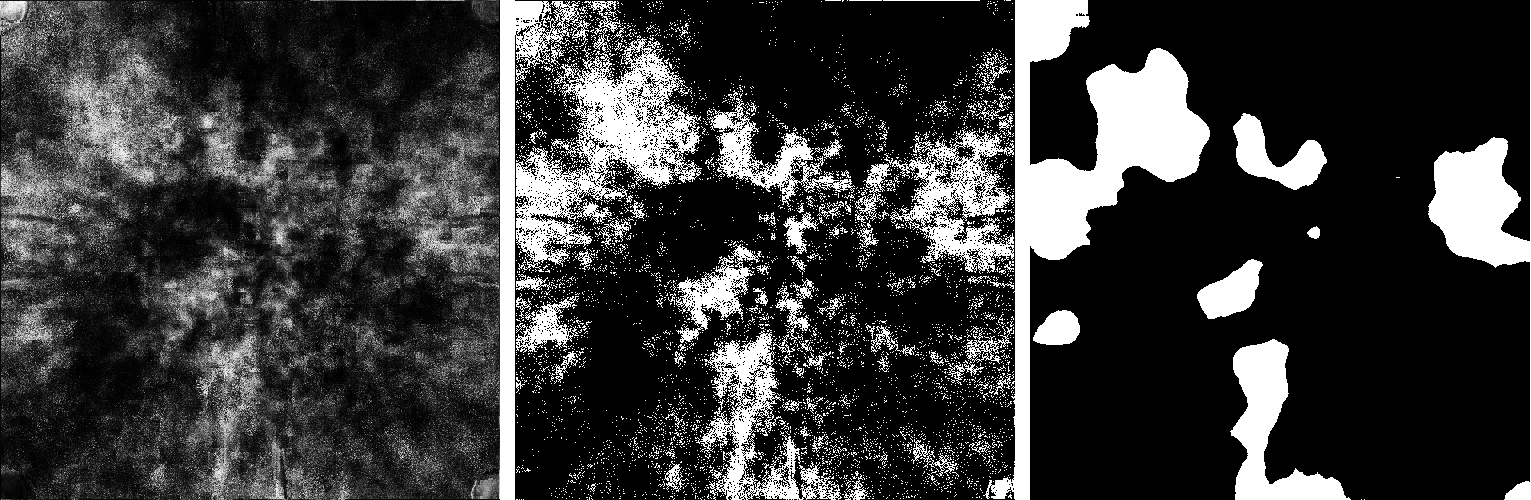}
    \caption{Starting from left, (a) a sample image which was generated using the trained generator of the GAN, (b) ground-truth estimate of the binary map (obtained using the unsupervised clustering algorithm), (c) smoothened binary map (to be finally used as the estimated ground-truth value).}
    \label{fig:GANimageGTestimation}
    \vspace{-0.5cm}
\end{figure}

Since the GAN only generates the sky/cloud images, we estimate the corresponding binary segmentation maps using an unsupervised clustering algorithm, as proposed by Dev.~et.~al.~\cite{dev2014systematic}. The clustering algorithm gives pixel-wise maps which were smoothened to estimate area-wise maps similar to the ones present in the SWINSEG dataset (see Fig.~\ref{fig:GANimageGTestimation} for reference results). We use the smoothened binary maps as the reference ground-truth maps for the images generated by GAN.

\subsection{PLS model to perform image segmentation}\label{sec:PLS}
PLS regression has already been used to perform cloud/sky image segmentation tasks for small image datasets~\cite{dev2017color}. Although more complex deep learning algorithms may perform better for this task, they also need huge datasets for effective training which is not the case here.

\begin{figure}[!ht]
    \centering
    \includegraphics[trim={0 0 0 65},clip,width=0.88\columnwidth]{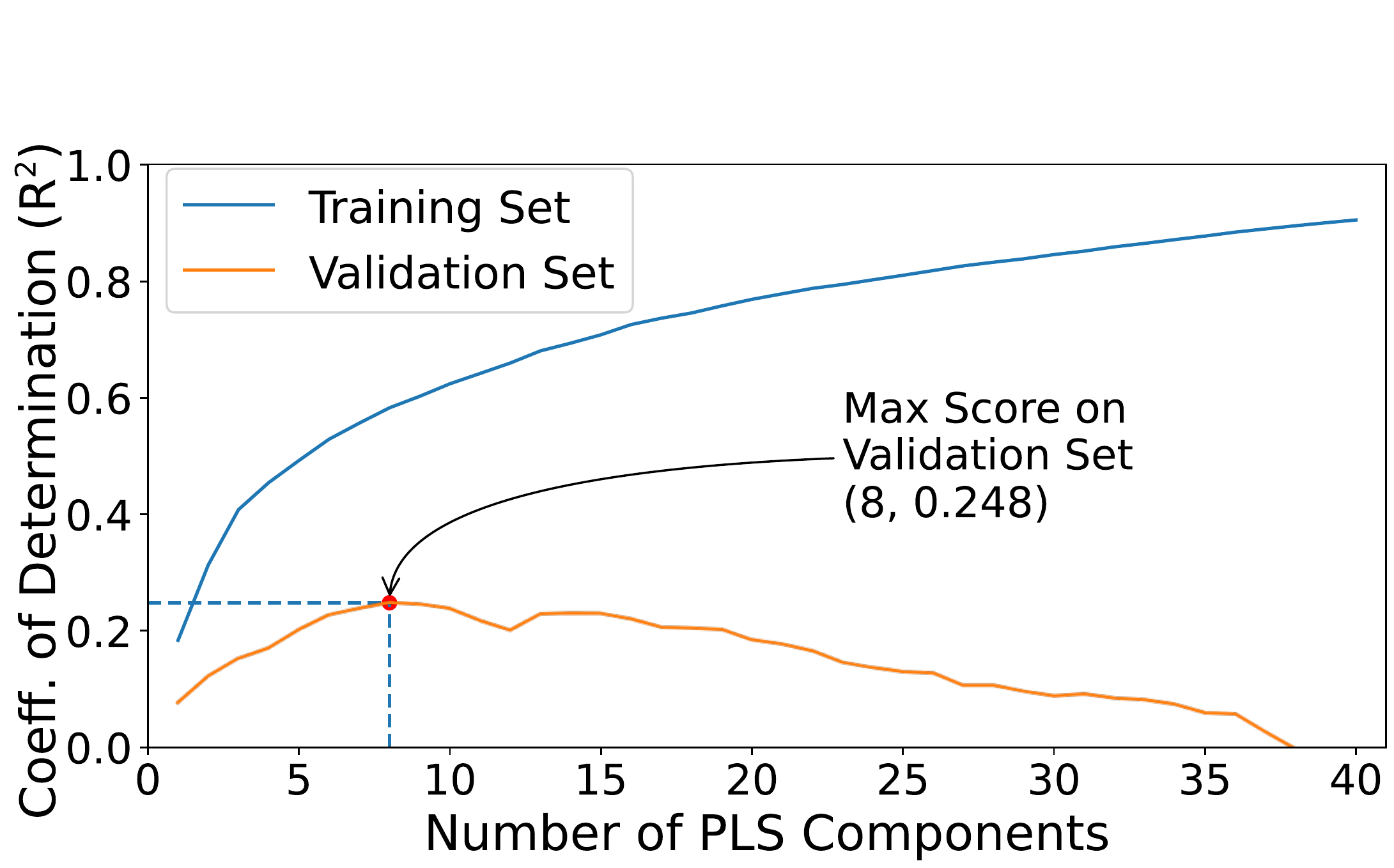}
    \caption{Hyper-parameter tuning to determine the optimal number of PLS components.}
    \label{fig:HPtuningPLS}
    \vspace{-0.5cm}
\end{figure}

This regression technique projects both the source and target variables to a lower dimension space before attempting to fit the regression model. The number of dimensions of the projected space is also called the `number of components' ($n\_comp$), which is a hyper-parameter for this technique. It is optimized by computing the coefficient of determination ($R^2$) for different values of $n\_comp$ on both training and validation sets. We pick the value of $n\_comp$ for which the value of $R^2$ on validation set is maximum, i.e. $n\_comp=8$ (see Fig.~\ref{fig:HPtuningPLS}).

\subsubsection{GAN Augmentation}
\label{sec:gan_augmentation}
Augmenting images generated from GAN is not straightforward due to the fact that the ground-truth maps were only estimated and not manually assigned. This leads to the fact that some GAN generated images and its corresponding binary maps (i.e. the estimated and smoothened ground-truth maps), or generated `data points', are highly incoherent or inaccurate. Such data points, when added to the training set act as huge outliers for the generic trend. Since any regression model doesn't ignore such outliers by default, they throw off the model from its original trajectory; thereby resulting in a low score (or high error rate). We removed any and all such outliers by utilizing the validation set once again.

\begin{figure}[!ht]
    \centering
    \includegraphics[width=0.88\columnwidth]{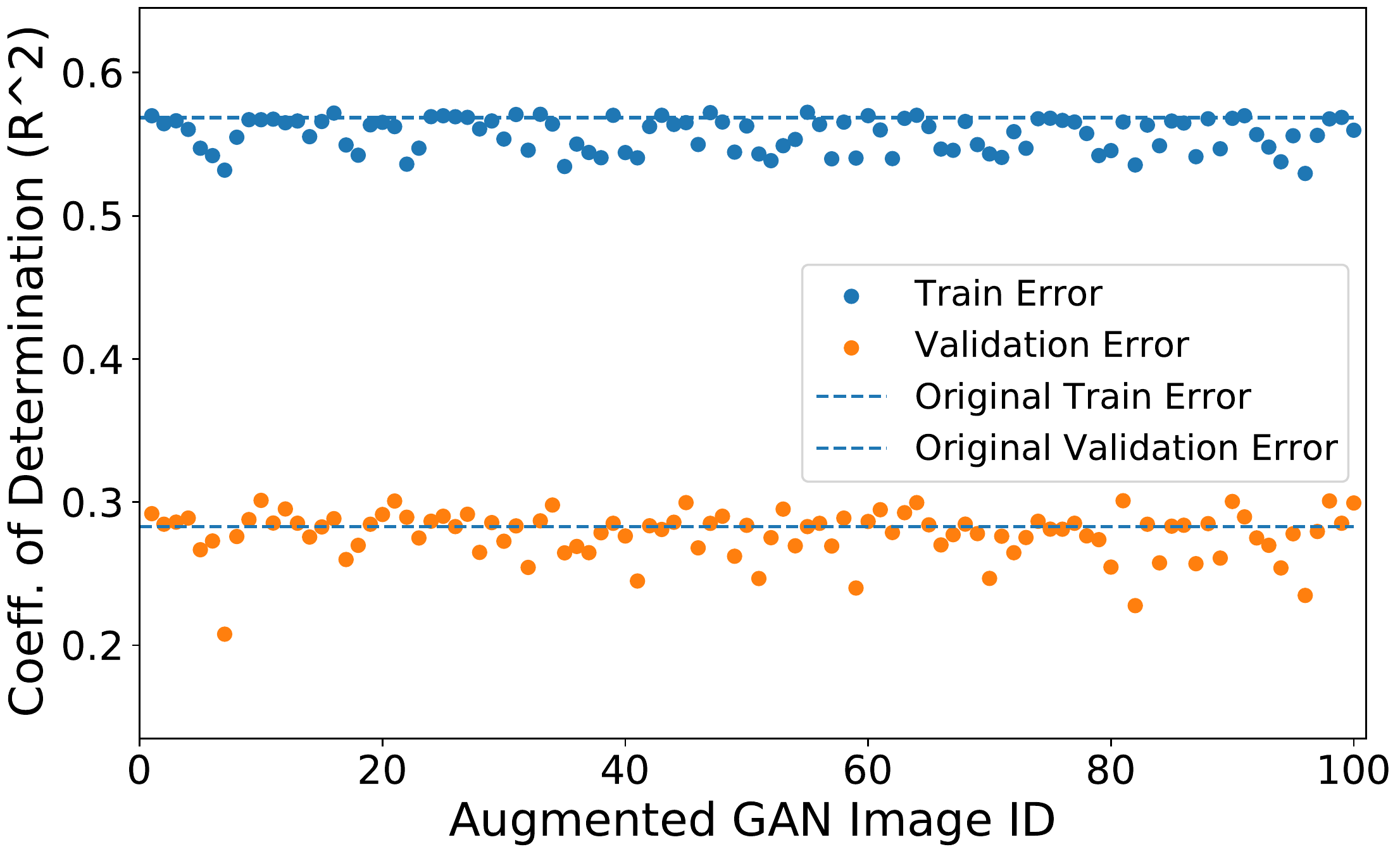}
    \caption{Finding favorable data points which were generated by the process described in Section~\ref{sec:GANimageGen}.}
    \label{fig:favourableGANimages}
    \vspace{-0.4cm}
\end{figure}

We augment the training set with one data point at a time and recalculate the value of $R^2$ on the training set and the validation set (see Fig.~\ref{fig:favourableGANimages}). If the newly obtained value of $R^2$ on validation set is less than the original value of $R^2$ on the validation set, the data point is declared as an outlier or unfavourable. Only the favourable data points were finally augmented to the training set on which final results are reported.

\section{Results}\label{sec:results}
To evaluate the effectiveness of augmenting training set with the GAN generated data points, we calculate the value of coefficient of determination ($R^{2}$) for the trained PLS model with and without augmentation. Table~\ref{table:R2scoreComparison} shows that post augmentation, $R^2$ decreases on training set but increases on test set. This shows that augmenting GAN generated images helps in better generalization of the model.

\begin{table}[htb!]
\small
\centering
\begin{tabular}{p{3.25cm}||C{1.9cm}|C{1.9cm}}
\hline
Cases & $R^2$ (Training) & $R^2$ (Test) \\ 
\hline\hline
Without Augmentation & $\bm{0.568}$ & $0.372$\\
After Augmentation & $0.539$ & $\bm{0.377}$\\
\hline
\end{tabular}
\caption{Coefficient of determination ($R^{2}$) as calculated when the PLS model was trained without augmenting the training set and after augmenting the training set.}
\label{table:R2scoreComparison}
\end{table}

\begin{figure}[!ht]
    \centering
    \begin{center}
        \includegraphics[width=0.88\columnwidth]{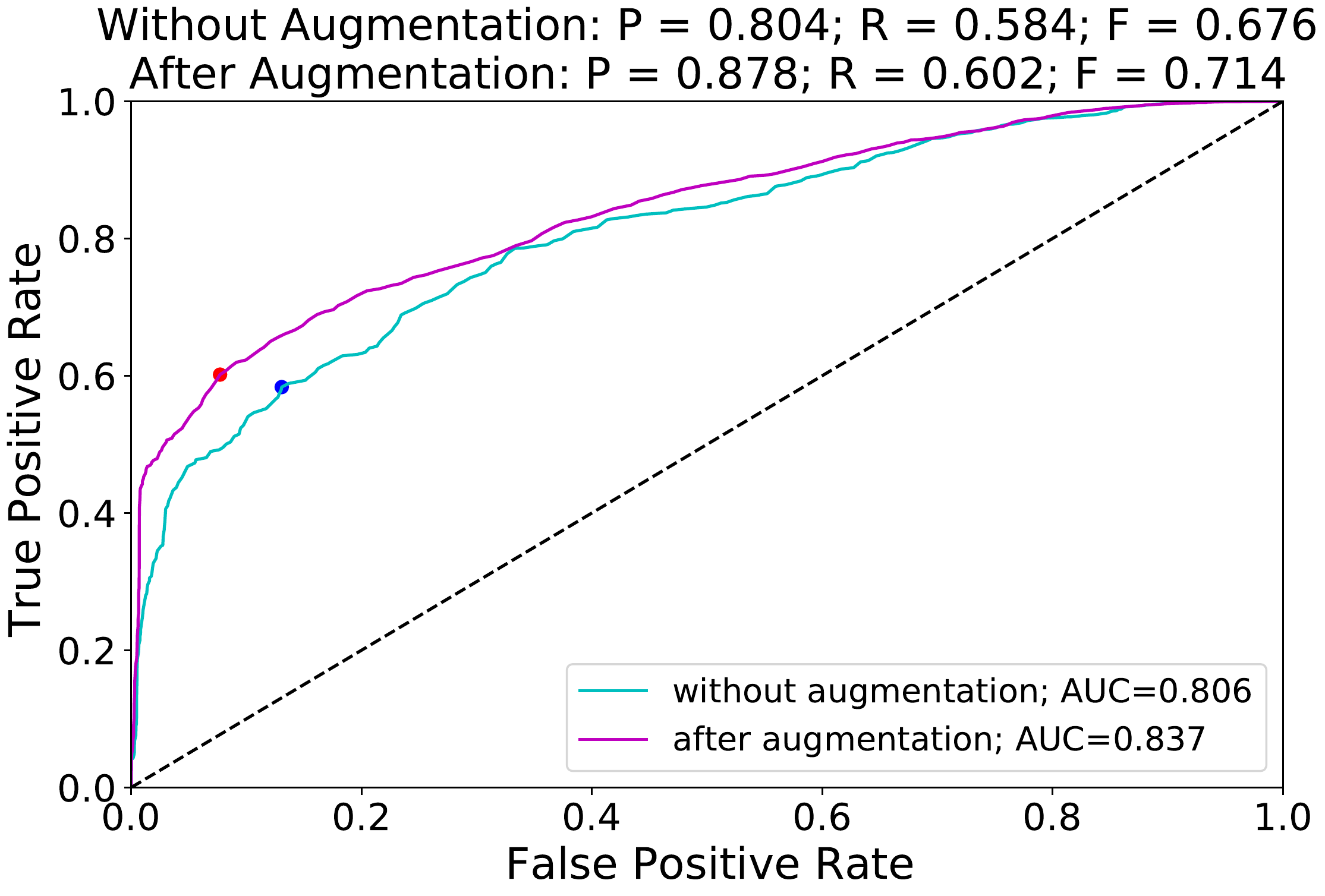}
        \vspace{-0.4cm}
    \end{center}
    \caption{ROC curves for a sample image from the test set as computed for without and with augmentation cases.}
    \label{fig:rocTestImages}
    \vspace{-0.35cm}
\end{figure}

Since the PLS model generates real numbered values in range $(-\infty, \infty)$, a threshold value (say, $thr$) needs to be identified to convert the predicted image into a binary segmentation map. For a particular value of $thr$, we created a confusion matrix for each image while considering clouds as positives and sky as negatives. The confusion matrix was used to plot the ROC (receiver operating characteristic) curves for each image and determine the optimal value of $thr$ (see Fig.~\ref{fig:rocTestImages}). Using the optimal values of $thr$, average values of \textit{Precision}, \textit{Recall} and \textit{F-Score} over the entire test set are computed. Table~\ref{table:PrecisionRecallFScore} shows an improvement in the \textit{F-Score} for the post augmentation case.

\begin{table}[htb!]
\small
\centering
\begin{tabular}{L{0.34\columnwidth}||C{0.145\columnwidth}|C{0.145\columnwidth}|C{0.145\columnwidth}}
\hline
Cases & Precision & Recall & F-Score \\ 
\hline\hline
Without Augmentation & $0.846$ & $\bm{0.749}$ & $0.776$\\
After Augmentation & $\bm{0.862}$ & $0.744$ & $\bm{0.781}$\\
\hline
\end{tabular}
\vspace{0.1cm}
\caption{Values of Precision, Recall and F-Score metrics when computed over the entire test set for the two cases.}
\label{table:PrecisionRecallFScore}
\end{table}
\vspace{-0.5cm}

\section{Conclusion \& Future Work}\label{sec:conclusion}
In this paper, we demonstrated the effectiveness of GANs in data augmentation for cloud/sky image segmentation tasks. While GANs were used for generating raw cloud/sky images, corresponding ground-truth binary segmentation maps were estimated using an unsupervised clustering algorithm. Even then we observed a slight improvement in both metrics, i.e. the coefficient of determination and the F-score, when the training set was augmented by GAN generated images. In future, we intend to modify the GAN architecture in order to generate ground-truth segmentation maps alongside the cloud/sky images. Further, we plan to verify their effectiveness when used alongside more complex prediction models like deep convolutional neural networks.

\vspace{-0.3cm}


\begin{thebibliography}{10}
\providecommand{\url}[1]{#1}
\csname url@samestyle\endcsname
\providecommand{\newblock}{\relax}
\providecommand{\bibinfo}[2]{#2}
\providecommand{\BIBentrySTDinterwordspacing}{\spaceskip=0pt\relax}
\providecommand{\BIBentryALTinterwordstretchfactor}{4}
\providecommand{\BIBentryALTinterwordspacing}{\spaceskip=\fontdimen2\font plus
\BIBentryALTinterwordstretchfactor\fontdimen3\font minus
  \fontdimen4\font\relax}
\providecommand{\BIBforeignlanguage}[2]{{%
\expandafter\ifx\csname l@#1\endcsname\relax
\typeout{** WARNING: IEEEtran.bst: No hyphenation pattern has been}%
\typeout{** loaded for the language `#1'. Using the pattern for}%
\typeout{** the default language instead.}%
\else
\language=\csname l@#1\endcsname
\fi
#2}}
\providecommand{\BIBdecl}{\relax}
\BIBdecl

\bibitem{ma2018application}
Z.~Ma, Q.~Liu, C.~Zhao, X.~Shen, Y.~Wang, J.~H. Jiang, Z.~Li, and Y.~Yung,
  ``Application and evaluation of an explicit prognostic cloud-cover scheme in
  grapes global forecast system,'' \emph{Journal of Advances in Modeling Earth
  Systems}, vol.~10, no.~3, pp. 652--667, 2018.

\bibitem{xiang2017very}
Z.~Xiang, W.~Ji, Z.~Hai, D.~Jie, C.~Fang, and Z.~Xin, ``Very short-term
  prediction model for photovoltaic power based on improving the total sky
  cloud image recognition,'' \emph{The Journal of Engineering}, vol. 2017,
  no.~13, pp. 1947--1952, 2017.

\bibitem{long2006retrieving}
C.~N. Long, J.~M. Sabburg, J.~Calb\'{o}, and D.~Pages, ``Retrieving cloud
  characteristics from ground-based daytime color all-sky images,''
  \emph{Journal of Atmospheric and Oceanic Technology}, vol.~23, no.~5, pp.
  633--652, 2006.

\bibitem{dev2017color}
S.~Dev, Y.~H. Lee, and S.~Winkler, ``Color-based segmentation of sky/cloud
  images from ground-based cameras,'' \emph{IEEE Journal of Selected Topics in
  Applied Earth Observations and Remote Sensing}, vol.~10, no.~1, pp. 231--242,
  2017.

\bibitem{sultana2020evolution}
F.~Sultana, A.~Sufian, and P.~Dutta, ``Evolution of image segmentation using
  deep convolutional neural network: A survey,'' \emph{Knowledge-Based
  Systems}, vol. 201-202, p. 106062, 2020.

\bibitem{dev2019cloudsegnet}
S.~Dev, A.~Nautiyal, Y.~H. Lee, and S.~Winkler, ``Cloudsegnet: A deep network
  for nychthemeron cloud image segmentation,'' \emph{IEEE Geoscience and Remote
  Sensing Letters}, vol.~16, no.~12, pp. 1814--1818, 2019.

\bibitem{dev2017nighttime}
S.~Dev, F.~M. Savoy, Y.~H. Lee, and S.~Winkler, ``Nighttime sky/cloud image
  segmentation,'' in \emph{IEEE International Conference on Image Processing
  (ICIP)}, 2017.

\bibitem{xie2020segcloud}
W.~Xie, D.~Liu, M.~Yang, S.~Chen, B.~Wang, Z.~Wang, Y.~Xia, Y.~Liu, Y.~Wang,
  and C.~Zhang, ``Segcloud: a novel cloud image segmentation model using deep
  convolutional neural network for ground-based all-sky-view camera
  observation,'' \emph{Atmospheric Measurement Techniques}, vol.~13, no.~4, pp.
  1953--1953, 2020.

\bibitem{goodfellow2014generative}
I.~Goodfellow, J.~Pouget-Abadie, M.~Mirza, B.~Xu, D.~Warde-Farley, S.~Ozair,
  A.~Courville, and Y.~Bengio, ``Generative adversarial nets,'' in
  \emph{Advances in neural information processing systems}, 2014, pp.
  2672--2680.

\bibitem{antoniou2017data}
A.~Antoniou, A.~Storkey, and H.~Edwards, ``Data augmentation generative
  adversarial networks,'' \emph{arXiv preprint arXiv:1711.04340}, 2017.

\bibitem{wegelin2000asurvey}
J.~A. Wegelin, ``A survey of partial least squares (pls) methods, with emphasis
  on the two-block case,'' Department of Statistics, University of Washington,
  Seattle, Tech. Rep., 2000.

\bibitem{kingma2014adam}
D.~P. Kingma and J.~Ba, ``Adam: A method for stochastic optimization,''
  \emph{arXiv preprint arXiv:1412.6980}, 2014.

\bibitem{dev2014systematic}
S.~Dev, Y.~H. Lee, and S.~Winkler, ``Systematic study of color spaces and
  components for the segmentation of sky/cloud images,'' in \emph{IEEE
  International Conference on Image Processing (ICIP)}, 2014, pp. 5102--5106.

\end{thebibliography}
\end{document}